\documentclass[10pt, a4paper]{article}

\usepackage{PMAN} 

\usepackage{booktabs}
\usepackage{multirow}
\usepackage{amssymb}
\usepackage{hyperref}

\hypersetup{
   breaklinks=true,   
   colorlinks=true,   
   pdfusetitle=true,  
}

\title{
Automatic Answerability Evaluation for Question Generation
}

\name{
Zifan Wang, Kotaro Funakoshi, Manabu Okumura
} 

\address{
Tokyo Institute of Technology \\
\{wangzf,funakoshi,oku\}@lr.pi.titech.ac.jp \\
}

\abstract{
Conventional automatic evaluation metrics, such as BLEU and ROUGE, developed for natural language generation (NLG) tasks, 
are based on measuring the n-gram overlap between the generated and reference text. 
These simple metrics may be insufficient for more complex tasks, such as question generation (QG), which requires 
generating questions that are answerable by the reference answers. 
Developing a more sophisticated automatic evaluation metric, thus, remains an urgent problem in QG research.
This work proposes \textbf{PMAN} (\textbf{P}rompting-based \textbf{M}etric on \textbf{AN}swerability), 
a novel automatic evaluation metric to assess whether the generated questions are answerable by the reference answers for the QG tasks.
Extensive experiments demonstrate that its evaluation results are reliable and align with human evaluations.
We further apply our metric to evaluate the performance of QG models, 
which shows that our metric complements conventional metrics. 
Our implementation of a GPT-based QG model achieves state-of-the-art performance in generating answerable questions.
\\ \newline  }

\begin{document}
\maketitleabstract
\section{Introduction}
Question Generation (QG) (\citealp{du-etal-2017-learning}; \citealp{yuan-etal-2017-machine}; \citealp{zhou2018neural}) and 
Multi-hop Question Generation (MQG) (\citealp{pan-etal-2020-semantic}; \citealp{su-etal-2020-multi}; \citealp{fei2022cqg}) are tasks of 
generating questions that are answerable by the specified reference answers from given passages.
Commonly used automatic evaluation metrics for QG tasks are 
BLEU \citep{papineni2002bleu}, ROUGE \citep{lin2004rouge}, and METEOR \citep{banerjee2005meteor}.
These metrics measure the n-gram overlap between the generated and reference text. 
However, it cannot assess whether the generated questions are answerable by the reference answers, though it is an essential requirement in the QG tasks. 
Consequently, previous work often supplements these metrics with human evaluations to assess the answerability.
However, human evaluations have apparent disadvantages: 
(1) They are time-consuming and expensive, and 
(2) Different work has conducted human evaluations under varying conditions, making it difficult to compare the performance of different QG models. 
Therefore, developing an automatic evaluation metric to assess answerability has become a crucial problem. 

To address this problem, we propose PMAN, Prompting-based Metric on ANswerability,
a novel automatic evaluation metric for the QG tasks, which assesses whether the generated questions are answerable with the supposed answers.
PMAN leverages the high 
instruction-following capabilities of recent large language models (LLMs) such as ChatGPT \citep{openai} and responds to input with "YES" or "NO" as its answerability.

We conducted extensive experiments to test the reliability of our proposed metric, 
using both manually created and model-generated samples. 
Our experiments demonstrate that the metric assessments are reliable and align with human evaluations. 
Furthermore, we applied our metric to evaluate the performance of QG models, 
showing that our metric complements conventional metrics. 
Additionally, we implemented a GPT-based QG model, 
which achieves state-of-the-art (SOTA) performance in generating answerable questions.

The main contributions of this work are two-fold: 
(1)~Proposing a novel automatic evaluation metric that measures answerability 
and conducting extensive experiments to validate its reliability; and
(2)~Demonstrating that our metric complements conventional metrics and a GPT-based MQG model achieves SOTA performance in generating answerable multi-hop questions.

\begin{figure*}[t]
\centering
\small
    \begin{tabularx}{\textwidth}{|c||X|}
        \hline
A        & Your task is to determine if the reference answer delimited by triple dashes is the answer to the question delimited by angle brackets, according to the passage delimited by triple backticks. \\
\cline{1-1}
C       & To solve the problem do the following: \\
        & - First, give your own answer to the question. \\
        & - Then compare your answer to the reference answer delimited by triple dashes and evaluate if the answer delimited by triple dashes is correct or not. Don't decide if the reference answer is correct until you have answered the question yourself. \\
\cline{1-1}
A        & - If the reference answer is correct, respond ``YES'', otherwise respond ``NO''. \\
        & \\
\cline{1-1}
B       & Passage: \verb|```|\textit{passage context}\verb|```| \\
        & Question: \verb|<|\textit{question context}\verb|>| \\
        & Reference answer: \verb|---|\textit{answer context}\verb|---|
        \\\hline
    \end{tabularx}
    \caption{CoT prompting for PMAN. Section A is the task description. B is the data. C is the CoT part.}
    \label{fig:Cot prompting}
\end{figure*}

\section{Related Work}
While automatic metrics measuring n-gram overlap are still commonly used for evaluating the performance of NLG models, 
they have been widely criticized for their low correlation with human judgments. 
Furthermore, these metrics cannot measure more nuanced properties, 
such as whether a generated question is answerable by a reference answer. 
To address these limitations, most work complements the automatic metrics with human evaluations (\citealp{xie-etal-2020-exploring}; \citealp{ji2021multi}), 
which are time-consuming and expensive.

The only significant work we could find that proposed a metric for measuring answerability is \citet{nema-khapra-2018-towards}.
They observed that specific components of a question are more relevant to its answerability, 
and the overlap of words in these components with words in the gold question should be assigned a higher score than the overlap of random words.
For example, if the gold question is ``Who was the director of Titanic?'', question A, ``director of Titanic?'' would be assigned a much lower score than question B, ``Who was the director of?'' by the existing overlap-based metrics. 
However, most humans would consider sentence A more answerable than B, 
as it contains a key component, a named entity ``Titanic.''
\citet{nema-khapra-2018-towards} identified four components (question types, content words, function words, and named entities) 
and proposed assigning a higher score for their overlap. 
Nevertheless, this approach is still an overlap-based metric, 
ignoring the one-to-many nature of the QG tasks.
As pointed out by \citet{yuan-etal-2017-machine} and \citet{ji2022qascore}, 
we can ask diverse questions that are answerable by the same answer.
For example, ``What person directed Titanic?'' would still have the same answer as the gold question ``Who was the director of Titanic?''
They are almost equally answerable. 
However, no overlap-based metric would assign a reasonable score to such questions.

Recently, new reference-free frameworks have been proposed to automatically evaluate NLG models. 
(\citealp{kocmi-federmann-2023-large}; \citealp{wang2023chatgpt}; \citealp{liu2023geval}; \citealp{mohammadshahi-etal-2023-rquge}). 
They employed large language models 
to assign scores for specific properties of generated texts, 
such as coherence and relevance.
This framework is 
reference-free such allows the generation of diverse questions, and 
can measure even more nuanced properties such as answerability, 
which were almost impossible to measure using the previous automatic metrics.
Especially, \citet{mohammadshahi-etal-2023-rquge} also proposed a metric for evaluating the quality of automatically generated questions, 
and explored the idea of leveraging the LLMs to answer the question. 
However, our work differs from theirs in the following aspects:
(1) Their QA model extracts a span in the input passage as the answer and thus cannot answer questions for which the answer does not appear in the passage (e.g., a question whose answer is either “yes” or “no”). Our model instead generates the answer to avoid this limitation; 
(2) Their metric requires an appropriate supervised training dataset in the target domain and language for the Span Scorer. Our approach does not;
(3) Their metric is not strictly intended to evaluate answerability. They describe their metric as “to compute the acceptability of the candidate”, which assesses the overall quality of generated questions, as a replacement for previous metrics. It mixes up multiple aspects including fluency and relevance. Therefore, when the assessed score is not high, it is not clear to the metric user whether it is due to fluency or answerability. Our work instead aims to assess whether the questions can be answered by the reference answer, as a complementary of previous metrics. To our knowledge, it is still the first work to effectively access the answerability.

\section{Proposed Metric}
\setlength{\tabcolsep}{2.5pt}
\begin{table*}[t]
\centering
\footnotesize
    \begin{tabular}{l l c| ccc | ccc | c | c}
        \toprule
           & Assessing &  & \multicolumn{3}{c|}{Answerable} & \multicolumn{3}{c|}{Non-answerable} &  & \# of valid \\
        Test sample set& model & CoT & Precision & Recall & F1 & Precision & Recall & F1 & Accuracy & responses \\
        \midrule
        \multirow{6}{*}{Manual (non-``yes/no'')}
        & GPT-4 & $\checkmark$ & \bf1.000 & 0.880 & 0.936 & 0.893 & \bf1.000 & 0.943 & 0.940 & 100 \\
        & GPT-4 &   & \bf1.000 & 0.820 & 0.901 & 0.847 & \bf1.000 & 0.917 & 0.910 & 100 \\
        & GPT-3.5 & $\checkmark$ & 0.976 & 0.800 & 0.879 & 0.831 & 0.980 & 0.899 & 0.890 & 100 \\
        & GPT-3.5 &  & \bf1.000 & 0.520 & 0.684 & 0.676 & \bf1.000 & 0.806 & 0.760 & 100 \\
        & Llama2 & $\checkmark$ & 0.880 & 0.880 & 0.880 & 0.880 & 0.880 & 0.88 &  0.880 & 100 \\
        & Llama2 &   & 0.980 & \bf0.980 & \bf0.980 & 0.980 & 0.980 & \bf0.980 & \bf0.980 & 100 \\
        \midrule
        \multirow{6}{*}{Manual (``yes/no'')}
        & GPT-4 & $\checkmark$ & 0.902 & \bf0.920 & 0.911 & \bf0.918 & 0.900 & 0.909 & 0.910 & 100 \\
        & GPT-4 &    & 0.978 & 0.880 & \bf0.926 & 0.891 & 0.980 & \bf0.933 & \bf0.930 & 100 \\
        & GPT-3.5 & $\checkmark$ & 0.623 & 0.760 & 0.685 & 0.692 & 0.540 & 0.607 & 0.650 & 100 \\
        & GPT-3.5 &              & \bf1.000 & 0.020 & 0.039 & 0.505 & \bf1.000 & 0.671 & 0.510 & 100 \\
        & Llama2* & $\checkmark$ & 0.870 & 0.833 & 0.851 & 0.790 & 0.833 & 0.811 & 0.833 & 48 \\
        & Llama2 &  & 0.806 & 0.580 & 0.674 & 0.672 & 0.860 & 0.754 & 0.720 & 100 \\
        \midrule
        Model (non-``yes/no'') 
        & GPT-4 & $\checkmark$ & 0.810 & 0.979 & 0.887 & 0.976 & 0.788 & 0.872 & 0.880 & 100 \\
        \midrule
        Model (``yes/no'') 
        & GPT-4 & $\checkmark$ & 0.851 & 0.966 & 0.905 & 0.939 & 0.756 & 0.838 & 0.880 & 100 \\
        \bottomrule
    \end{tabular}
    \caption{Reliability evaluation of PMAN (Llama2* was evaluated only on 48 valid responses). 
    Manually created samples are balanced: half of the questions are answerable, while model-generated samples unbalanced data: 48 answerable vs. 52 non-answerable for non-``yes/no'', 59 vs. 41 for ``yes/no''.
    }
    \label{tab:reliability}
\end{table*}

This section presents the design and computation of our proposed evaluation metric {PMAN},
the \ul{P}rompting-based \ul{M}etric on \ul{AN}swerability.
The fundamental task of this metric is 
to determine if the reference answer is an appropriate answer to a generated question 
in accordance with the presupposed passage.

\subsection{Prompt Design}
Standard prompting describes the target task.
\citet{wei2022chain} discovered that 
providing a Chain-of-Thought (CoT) outlining the steps to perform a reasoning task would significantly improve LLMs' performance. 
Thus, we investigate the effectiveness of CoT for our evaluation task.
For our case, we provide step-by-step instructions asking a GPT-based LLM to:
(1)~answer a question itself; 
(2)~compare its answer to the reference answer; and 
(3)~give the final assessment by responding with ``YES'' or ``NO.''
Figure \ref{fig:Cot prompting} presents the framework of our CoT prompting, consisting of three sections.
When we adopt CoT, we use all sections.
When not, we exclude section C from the prompt.

\subsection{Metric Computation}
We initially set the decoding temperature parameter to 0.
For each question, we instruct an LLM to give an assessment.
The assessment is valid if it contains ``YES'' or ``NO.''
If not,
we increase the decoding temperature and request the model to regenerate a response until it validly assesses the question.
Consequently, each question is likely associated with a valid assessment containing ``YES'' or ``NO.''
Then, we calculate the PMAN score as the percentage of questions with ``YES'' assessments among all assessed questions.
\section{Reliability of the Metric}
We experimentally evaluated the reliability of PMAN with manually created and model-generated samples.

\subsection{Manually Created Test Samples}
We manually created 
the test samples
using 
the HotpotQA \citep{yang-etal-2018-hotpotqa} dataset. 
A sample consists of a passage and a gold question-answer pair on it.
We exactly used the randomly selected samples in the dataset for answerable questions.
We used forged samples for non-answerable questions, in which the original question and answer are replaced with those from another randomly chosen sample.

Although ``yes/no''-type questions cover only six percent of HotpotQA~\citep{yang-etal-2018-hotpotqa},
in preliminary investigations, we observed that LLMs tend to perform quite differently on this type of question.
Therefore, we created 100 samples for both ``yes/no'' and non-``yes/no'' types and conducted separate tests with them.
We evaluated manually created samples using the following three different LLMs, i.e., GPT-4 (gpt-4-0613), GPT-3.5 (gpt-3.5-turbo-0613), and Llama2 (llama-2-70b-chat),
with two variations of \textit{with} and \textit{without} CoT for each.







According to the results shown in the upper half of Table \ref{tab:reliability}, 
(1)~PMAN can demonstrate an acceptable performance, especially with 
a reliability of more than 90\% when using GPT-4;
(2)~
CoT prompting is effective for GPT-3.5 but not so much for the other two models; and
(3) 
PMAN can be used with
different LLMs. 
However, in the case of Llama2, nearly 60 percent of responses to ``yes/no'' type questions were invalid 
, i.e., could not provide an expected response, even after multiple re-generations.

\subsection{Model Generated Test Samples}

\begin{figure*}[t]
\centering
\small
    \begin{tabularx}{\textwidth}{|X|}
        \hline
        Generate a question from the passage delimited by triple backticks that can be answered by the answer delimited by triple dashes, where answering the question requires reasoning over multiple sentences in the passage.
        \\\\
        Passage: \verb|```|passage contexts\verb|```|
        \\
        Answer: \verb|---|question contexts\verb|---|
        \\
        Question: 
        \\\hline
    \end{tabularx}
    \caption{Prompt used in GPT-based question generation}
    \label{fig:ChatGPT}
\end{figure*}

The high capability of
assessing manually created high-quality samples cannot necessarily guarantee alignment with humans for model-generated questions. 
Low-quality 
questions could fool a metric.

Thus, we further tested PMAN with 100 questions generated by EQG \citep{su-etal-2020-multi}, SQG \citep{pan-etal-2020-semantic}, CQG \citep{fei2022cqg}, and
GPT-4 with the prompt shown in Figure~\ref{fig:ChatGPT}.
We generated 100 questions for "yes/no" and non-"yes/no" types
using pairs of passages and corresponding reference answers randomly selected from the HotpotQA dataset. 
The answerability labels of generated questions were determined by the authors.
For the non-``yes/no'' type, each model provided 25 questions.
For the ``yes/no'' type, SQG provided 50 instead of EQG, as it did not handle the ``yes/no'' type.
The results in the lower half of Table \ref{tab:reliability} demonstrate a strong alignment between PMAN using GPT-4 with CoT and human evaluations. 

\begin{table}[t]
\centering
\tabcolsep=2.05pt
\footnotesize
    \begin{tabular}{c |c c c c c c |c c}
        \toprule
        Model & BL-1 & BL-2 & BL-3 & BL-4 & RG-L & MTR & PMAN  & HM\\
        \midrule
        EQG & 40.2 & 26.7 & 19.7 & 15.2 & 35.3 & 20.5 & 42 & 40\\
        SQG & 40.7 & 27.3 & 20.1 & 15.4 & 36.9 & 20.3 & 41 & 33\\
        CQG & 50.8 & 37.8 & 30.2 & 24.8 & 46.0  & 25.1 & 76 & 69\\
        GQG & 38.8 & 25.4 & 18.5 & 14.3 & 35.1 & 21.9 & 97 & 97\\
        \bottomrule
    \end{tabular}
    \caption{Evaluation results.     
    {BL, RG, and MTR stand for BLUE, ROUGE, and METEOR, respectively.
    HM is human assessment of answerability.}   
 }
    \label{tab:Evaluation Results}
\end{table}

\begin{table}
\centering
\footnotesize
    \begin{tabularx}{\linewidth}{c|X}
    \toprule
    Passage & 
    Heinrich August Marschner (16 August 1795 - 14 December 1861) was the most important composer of German opera between Weber and Wagner. Carl Maria Friedrich Ernst von Weber (18 or 19 November 1786 - 5 June 1826) was a German composer, conductor, pianist, guitarist and critic, and was one of the first significant composers of the Romantic school.\\
    \midrule
    Ref.\,Q. & Heinrich Marschner was a composer who performed in the time frame after one of the first significant composers in what school of work? 
    (Answer: Romantic)\\
    \midrule
    EQG & heinrich august marschner was the most important composer of german opera between weber and a german composer who was one of the first significant composers of what school ?\\
    SQG & what genre did both heinrich marschner and karl von weber share ?\\
    CQG & heinrich marschner and carl maria von weber both were composers of what school ?\\
    GQG & What school of music was Carl Maria Friedrich Ernst von Weber, a significant German composer who came before Heinrich August Marschner, associated with?\\
    \bottomrule
    \end{tabularx}
\caption{Examples of generated questions.}\label{tab:examples}
\end{table}

\section{Application of the Metric}
\label{sec:evaluations}

We applied PMAN to the MQG models' output along with other metrics. 
An MQG model we implemented with GPT-4 achieved SOTA in generating answerable questions.

\subsection{Dataset}
We used the HotpotQA dataset \citep{yang-etal-2018-hotpotqa}. 
The size of the test set 
is 6,972.
Answering each question in the dataset necessitates the ability to reason over supporting sentences merged in a passage from two distinct Wikipedia documents. 

\subsection{Models}
The aforementioned three conventional models
and a MQG model using GPT-4 were employed.
EQG~\citep{su-etal-2020-multi} 
and SQG~\citep{pan-etal-2020-semantic} 
were among the earliest work 
for the MQG task, 
while CQG~\citep{fei2022cqg} 
achieved the SOTA. 
We also implemented GQG, a MQG model using OpenAI GPT-4 (gpt-4-0613) with the prompt shown in Figure \ref{fig:ChatGPT}.





\subsection{Metrics}
In addition to PMAN,  three conventional metrics (BLEU \citep{papineni2002bleu}, ROUGE \citep{lin2004rouge}, and METEOR \citep{banerjee2005meteor}) were applied. 
PMAN was implemented on gpt-4-0613 using the prompt shown in Figure \ref{fig:Cot prompting}. 
All PMAN scores were calculated using the same 100 samples randomly selected from the test set.\footnote{These 100 samples are different from the 100 samples used in section 4} In contrast, the other metrics were calculated using the whole test set except for GQG, which was evaluated using the 100 samples mentioned above.
The answerability of generated questions was also assessed by the authors.
    
\subsection{Results}
Table \ref{tab:Evaluation Results} shows the results.
The rank agreement between PMAN (automatic assessment) and HM (human assessment) is perfect.

Concerning EQG, SQG, and GQG, there is a seemingly fair correlation between conventional metrics and PMAN. 
However, while BLUE and ROUGE evaluate GQG as the worst, 
GQG achieved the best answerability.
We think this is because, as exemplified in Table~\ref{tab:examples}, there is a significant qualitative leap between GQG and the other MQG models, 
which goes beyond the capability of the overlap-based metrics.
Therefore, PMAN could be complementary to them or even replace them.
Last but not least, as our implemented GQG model achieves the highest PMAN score,
it would serve as a new strong baseline model for generating answerable questions.

\section{Conclusion}
This paper highlighted an urgent issue in question generation: 
the absence of an effective automatic evaluation metric to assess whether the generated questions are answerable.
We proposed a Prompting-based Metric on ANswerability (PMAN) to address this issue, leveraging recent LLMs. 
Experiments with manually created and model-generated samples demonstrated reliability and strong alignment with human evaluations. 
The experiments also suggested that the notable Chain-of-Thought (CoT) prompting might not be effective for the most advanced models in some tasks.
A comparison of our metric to other conventional metrics further indicated its potential to complement the conventional metrics and to guide future research in QG toward the generation of more answerable questions.

\section{References}
\bibliographystyle{PMAN}
\bibliography{PMAN}

\end{document}